\documentclass[letterpaper]{article} 
\usepackage{aaai25}  
\usepackage{times}  
\usepackage{helvet}  
\usepackage{courier}  
\usepackage[hyphens]{url}  
\usepackage{graphicx} 
\usepackage{natbib}  
\usepackage{caption} 
\usepackage{algorithm}
\usepackage{algorithmic}

\usepackage{amsmath,amsfonts,amssymb,bm}
\usepackage{bibentry}
\usepackage{multirow}
\usepackage{subcaption}
\usepackage{xspace}

\makeatletter
\DeclareRobustCommand\onedot{\futurelet\@let@token\@onedot}
\def\@onedot{\ifx\@let@token.\else.\null\fi\xspace}

\def\eg{\emph{e.g}\onedot} 
\def\ie{\emph{i.e}\onedot}

\makeatother

\usepackage{newfloat}
\usepackage{listings}
\DeclareCaptionStyle{ruled}{labelfont=normalfont,labelsep=colon,strut=off} 
\floatstyle{ruled}
\newfloat{listing}{tb}{lst}{}
\floatname{listing}{Listing}
\title{DreamPhysics: Learning Physics-Based 3D Dynamics with Video Diffusion Priors}
\author{
    Tianyu Huang$^{1,2}$ \ \ \ \ Haoze Zhang$^{1}$ \ \ \ \ Yihan Zeng$^{3}$ \ \ \ \ Zhilu Zhang$^{1}$ \\ Hui Li$^{1}$ \ \ \ \ Wangmeng Zuo$^{1,}$\thanks{Joint corresponding authors.} \ \ \ \ Rynson W. H. Lau$^{2,}$\footnotemark[1]
}
\affiliations{
    \textsuperscript{1} Harbin Institute of Technology \\
    \textsuperscript{2} City University of Hong Kong \\  
    \textsuperscript{3} Huawei Noah's Ark Lab
}

\usepackage{bibentry}

\begin{document}

\maketitle

\begin{figure*}[t]
   \centering
   \includegraphics[width=0.85\textwidth]{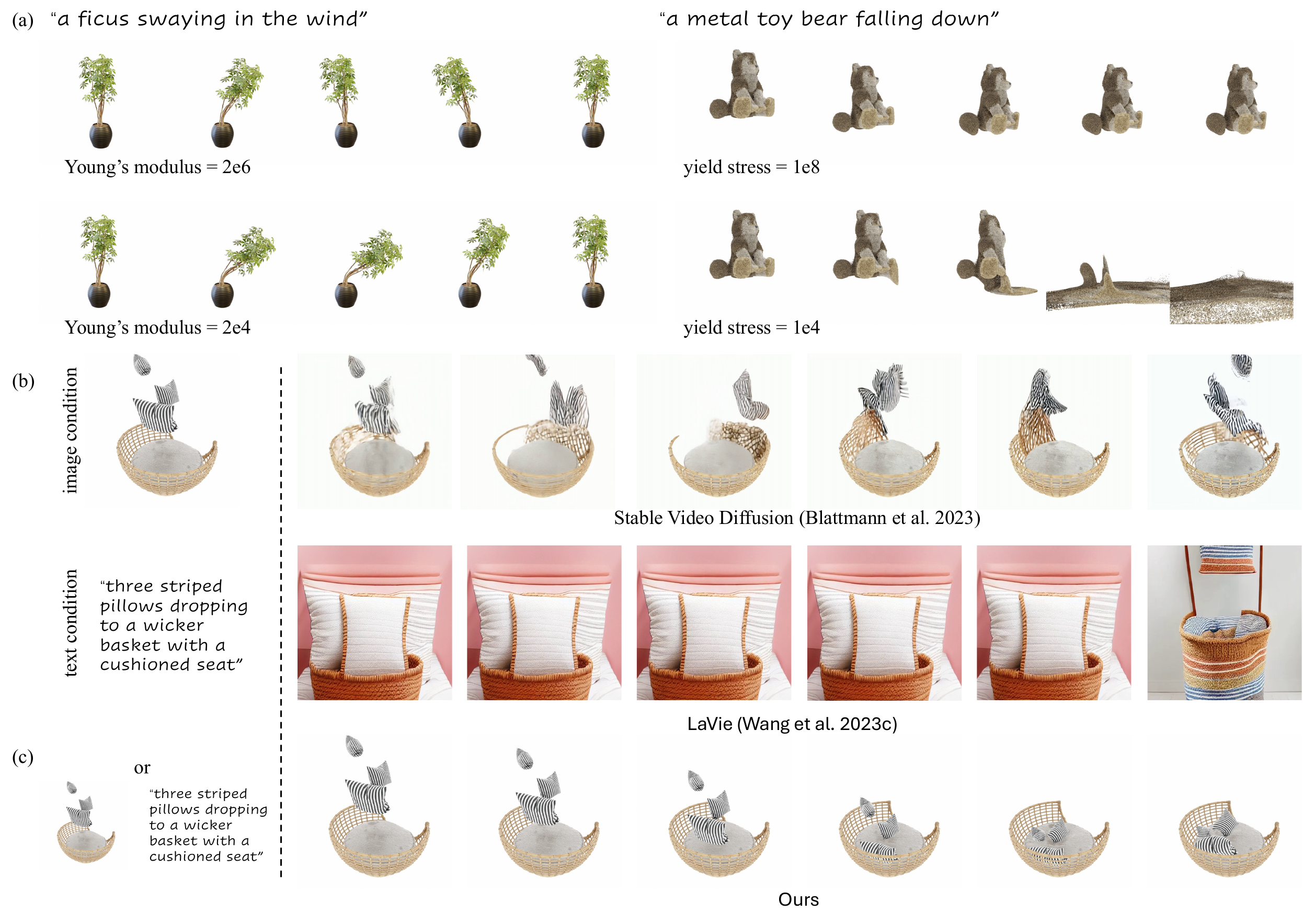}
   \caption{(a): The setting of physical properties can significantly affect the quality of the simulated videos. (b) Using state-of-the-art video diffusion models~\cite{blattmann2023stable,wang2023lavie} can hardly generate the desired results. (c) Our DreamPhysics can produce realistic 3D dynamic content with the distillation of video diffusion priors.}
   \label{fig:teaser} 
\end{figure*}

\begin{abstract}
Dynamic 3D interaction has been attracting a lot of attention recently. However, creating such 4D content remains challenging. One solution is to animate 3D scenes with physics-based simulation, which requires manually assigning precise physical properties to the object or the simulated results would become unnatural. Another solution is to learn the deformation of 3D objects with the distillation of video generative models, which, however, tends to produce 3D videos with small and discontinuous motions due to the inappropriate extraction and application of physics priors. In this work, to combine the strengths and complementing shortcomings of the above two solutions, we propose to learn the physical properties of a material field with video diffusion priors, and then utilize a physics-based Material-Point-Method (MPM) simulator to generate 4D content with realistic motions. In particular, we propose motion distillation sampling to emphasize video motion information during distillation. In addition, to facilitate the optimization, we further propose a KAN-based material field with frame boosting. Experimental results demonstrate that our method enjoys more realistic motions than state-of-the-arts do.
\end{abstract}

\begin{links}
    \link{Code} {https://github.com/tyhuang0428/DreamPhysics}
\end{links}

\section{Introduction}

With the development in 3D representations, \eg, Neural Radiance Fields (NeRF)~\cite{mildenhall2021nerf} and 3D Gaussian Splatting (GS)~\cite{kerbl3Dgaussians}, significant progress has been made in creating 3D assets through reconstruction and generation~\cite{poole2022dreamfusion,wang2024prolificdreamer}.
However, interacting with these 3D assets in a simulation environment~\cite{savva2019habitat,xia2018gibson} remains challenging, despite its importance in many applications, \textit{e.g.}, video games~\cite{fan2022minedojo}, virtual reality~\cite{jiang2024vr}, and robotics~\cite{lu2024manigaussian}.

Animating static 3D objects based on instructions is an important step toward this interaction goal.
In the real world, object movement is intertwined with the object's internal properties (\eg, material types).
Hence, we can see that on the one hand, some works~\cite{xie2023physgaussian,feng2024gaussian} first inject physical parameters into 3D GS objects, and then perform motion predictions in a physics-based simulator. 
However, as all these parameters have to be manually assigned, it is difficult to set them accurately, thus producing unnatural simulation results, as demonstrated in Figure~\ref{fig:teaser}(a). 
On the other hand, pre-trained video generators~\cite{singer2022make,khachatryan2023text2video,wang2023modelscope} are trained on real-world video data, which has naturally incorporated physical phenomena and regulations. 
These generators should contain, to some extent, physics-based prior knowledge. 
Thus, some works~\cite{singer2023text,bahmani20234d,zhao2023animate124} directly learn time-dependent deformation with the distillation of video diffusion models. 
However, the generated motions tend to exhibit small and discontinuous motions across frames.
We hypothesize that the main reason for this drawback is the inappropriate extraction and application of the physics prior, rather than the utilization of video models. 
We therefore ask this question: how can we mine and apply the physics knowledge of video generative models to achieve realistic dynamic 3D synthesis?

To this end, we rethink the usage of physics-based simulation and video generative models in this work. We propose to learn a material field, rather than a deformation field, from video diffusion models, and then deploy a physics-based simulator to animate the 3D object in this field.
As such, the advantages of the above two related approaches are combined, while their shortcomings can be complemented.
Learnable physical properties from video diffusion models eliminate the need for manual modulation, and the physics simulator based on reasonable properties ensures more realistic motion generation.

Specifically, we introduce a new framework named DreamPhysics.
DreamPhysics takes 3D GS~\cite{kerbl3Dgaussians} as a 3D representation.
It first learns the physical properties of a material field with the distillation of video diffusion priors, and then adopts a simulator based on Material Point Method (MPM)~\cite{stomakhin2013material,jiang2016material} to model the time-dependent deformation of each Gaussian kernel. 
During the distillation from video diffusion models, the Score Distillation Sampling (SDS)~\cite{poole2022dreamfusion} may focus more on color information, and is not completely suitable for extracting motion information. Instead, we propose motion distillation sampling (MDS) to avoid the interference of color bias and emphasize the motion information in the rendered video.
In addition, directly optimizing the material field can easily lead to unstable training due to the large range of possible parameter values. To facilitate the training process, we propose a KAN-based~\cite{liu2024kan} material field with frame boosting.

We note that there is a concurrent work named PhysDreamer~\cite{zhang2024physdreamer}, which supervises the prediction of physical properties with a ground-truth video generated by an image-to-video diffusion model. However, as shown in Figure~\ref{fig:teaser}(b), the video generative model can hardly produce the desired results to serve as ground truth, due to its poor motion control over the image/text condition. 
In contrast, our DreamPhysics supports both image-conditioned and text-conditioned optimization without the need for pre-generated ground truth, as demonstrated in Figure~\ref{fig:teaser}(c).
Experimental results demonstrate that our method can effectively distill the video diffusion prior and assign proper values to the physical properties. Compared with state-of-the-art works, our results enjoy more realistic motion.

Our main contributions can be summarized as:
\begin{itemize}
    \item We introduce a physics-based 3D animation framework, \ie, DreamPhysics, which learns a material field for a physics simulator to support the creation of dynamic 3D content.
    \item We propose motion distillation sampling for the optimization of physical properties with video diffusion priors. To facilitate the optimization, we further propose a KAN-based material field with frame boosting.
    \item DreamPhysics can generate high-quality 4D content with either image- or text-conditioned optimization. Extensive experiments show that our results enjoy more realistic motion simulation.
\end{itemize}

\section{Related Work}
\subsection{3D Generation}
In recent years, 3D generation has advanced significantly, with methods broadly classified into two main categories: 3D supervised and 2D lifting approaches.

3D supervised methods~\cite{nichol2022point,jun2023shap,yu2023towards,huang2023textfield3d,hong2023lrm,tang2024lgm} utilize text-3D data to train generators capable of directly producing 3D assets. For instance, Point-E~\cite{nichol2022point} is an early example of a text-to-3D generator that creates point clouds based on input prompts. Shap-E~\cite{jun2023shap} and LGM~\cite{tang2024lgm} have expanded the scope of generated content to include SDF~\cite{park2019deepsdf} and 3DGS~\cite{kerbl3Dgaussians} representations, respectively. Despite their efficiency in generating solid 3D content, these methods are significantly limited by the availability of 3D data. The current scale of 3D training datasets~\cite{reizenstein2021common,deitke2023objaverse,deitke2024objaverse} is much smaller compared to 2D or video datasets, resulting in a constrained open-world capability relative to image or video generators. TextField3D~\cite{huang2023textfield3d} attempts to enhance text control in 3D generators using a noisy latent space, yet it still falls short of achieving the imaginative capabilities seen in 2D generators.

Conversely, 2D lifting methods~\cite{poole2022dreamfusion,lin2023magic3d,metzer2023latent,Chen_2023_ICCV,wang2024prolificdreamer} leverage the extensive prior knowledge embedded in 2D diffusion models to optimize 3D representations. DreamFusion~\cite{poole2022dreamfusion} pioneered the concept of score distillation sampling (SDS), which distills 3D renderings into 2D diffusion. Although these methods produce photorealistic results, they are prone to 3D inconsistency issues, commonly referred to as the Janus problem.

To address this issue, recent works~\cite{liu2023zero,shi2023mvdream,long2023wonder3d} have explored the synthesis of multi-view images of 3D objects. For example, Zero-1-to-3~\cite{liu2023zero} generates images of the same object from different viewpoints based on a given image and viewpoint angles. MVDream~\cite{shi2023mvdream} enhances consistency by generating orthogonal multi-view images of the same object. Wonder3D~\cite{long2023wonder3d} supports depth generation to achieve more precise object reconstruction.

In this work, we collect static 3D scenes from both reconstruction data and 3D generation methods, providing more available assets for evaluation.

\subsection{3D Animation}
3D animation creation has significantly increased demand across various applications, such as video games, virtual reality, and robotic simulation. However, manually creating such 4D content is a time-consuming process that necessitates a high level of expertise. To animate a 3D object, the common practice is to bind the object with a template skeleton, also known as rigging. TADA~\cite{liao2023tada} produces 3D assets based on SMPL-X~\cite{pavlakos2019expressive}, which is a human-body 3D template that supports animation. DreamControl~\cite{huang2024dreamcontrol} proposes to generate 3D assets conditioned by input skeletons, which can be rigged easily for animation.

As the success of video generative models~\cite{wang2023lavie,wang2023modelscope,blattmann2023stable,zhang2023controlvideo}, some methods~\cite{zhao2023animate124} attempt to leverage video diffusion models to guide the prediction of the 3D deformation. DreamGuassian4D~\cite{ren2023dreamgaussian4d} uses a pre-generated video to supervise the deformation of static scenes. Animate124~\cite{zhao2023animate124} proposes to distill the priors of video diffusion models to its deformation fields.

The deformation prediction in these methods is not accurate. Recent works~\cite{xie2023physgaussian,feng2024splashing,zhang2024physdreamer} introduce physics simulation to the 3D deformation. PhysGaussian~\cite{xie2023physgaussian} deploys the finite element method to model the deformation of elastic objects like collision and shaking. \citeauthor{feng2024splashing} further supports the simulation of liquid. However, these methods require manually setting the physical properties for objects before simulation. PhysDreamer~\cite{zhang2024physdreamer} attempts to optimize these properties with pre-generated videos, but the quality of generated videos can hardly be ensured. In this work, we propose to distill the priors of video diffusion models to simulation environments, enabling the automatic setting of physical properties.

\section{Preliminaries}
\subsection{Point-Based Representation}
Point cloud~\cite{guo2020deep} is an explicit 3D representation, which generally consists of the coordinates for all points. Normal and color information~\cite{dai2017scannet,qi2017pointnet} can also be considered to further enrich the feature space of the point cloud. Despite the succinct representation, its rendering quality is heavily restricted by the number of points~\cite{huang2023clip2point}. Derived from NeRF~\cite{mildenhall2021nerf}, 3D Gaussian Splatting (GS)~\cite{kerbl3Dgaussians} introduces a point-based explicit radiance field. Points are modeled as a set of Gaussian kernels $\{\mathcal{G}_i\} = \{x_i, \sigma_i, \Sigma_i, C_i\}$, where $x_i$, $\sigma_i$, $\Sigma_i$, and $C_i$ denote the center coordinate, opacity, covariance matrix, and spherical harmonic coefficient of the $i$-th kernel $\mathcal{G}_i$. To render a 3D GS scene at a specific viewpoint $\mathbf{r}$, the color can be formulated as:
\begin{equation}
\label{eq:3dgs}
   \mathbf{C} = \sum_{i=1}^N T_i \alpha_i C_i, \ \rm{with}\; T_i=\prod_{j=1}^{i-1} (1-\alpha_j),
\end{equation}
where $N$ is the set of sorted Gaussian kernels related to the pixel and the viewpoint. $\alpha_i$ is the effective opacity given by evaluating a 2D Gaussian with $\Sigma$ and $\sigma$. 3D GS can reconstruct high-fidelity views by real-time rendering, and support explicit interaction and editing.

\subsection{Material Point Method}
The material point method (MPM)~\cite{stomakhin2013material,jiang2016material} is a numerical simulation mechanic for the analysis of continuum forces. In MPM, the continuum is represented by a set of particles placed in a grid-based space. Different from mesh-based numerical mechanics, MPM can be naturally applied to point-based representation 3D GS. Following PhysGaussian~\cite{xie2023physgaussian}, we have a time-dependent state for each Gaussian kernel as:
\begin{equation}
\label{eq:mpm}
    x_i(t) = \Delta (x_i, t), \ \Sigma_i(t) = F_i(t) \Sigma_i F_i(t)^T,
\end{equation}
where $\Delta(\cdot,t)$ and $F_i(t)$ are the coordinate deformation and the deformation gradient at timestep $t$. Considering the continuum rotation $\Omega_i(t)$, the rendering viewpoint also requires adjustment to satisfy the view direction of spherical harmonic coefficient $C_i$.

\begin{figure*}[t]
  \centering
  \includegraphics[width=0.85\textwidth]{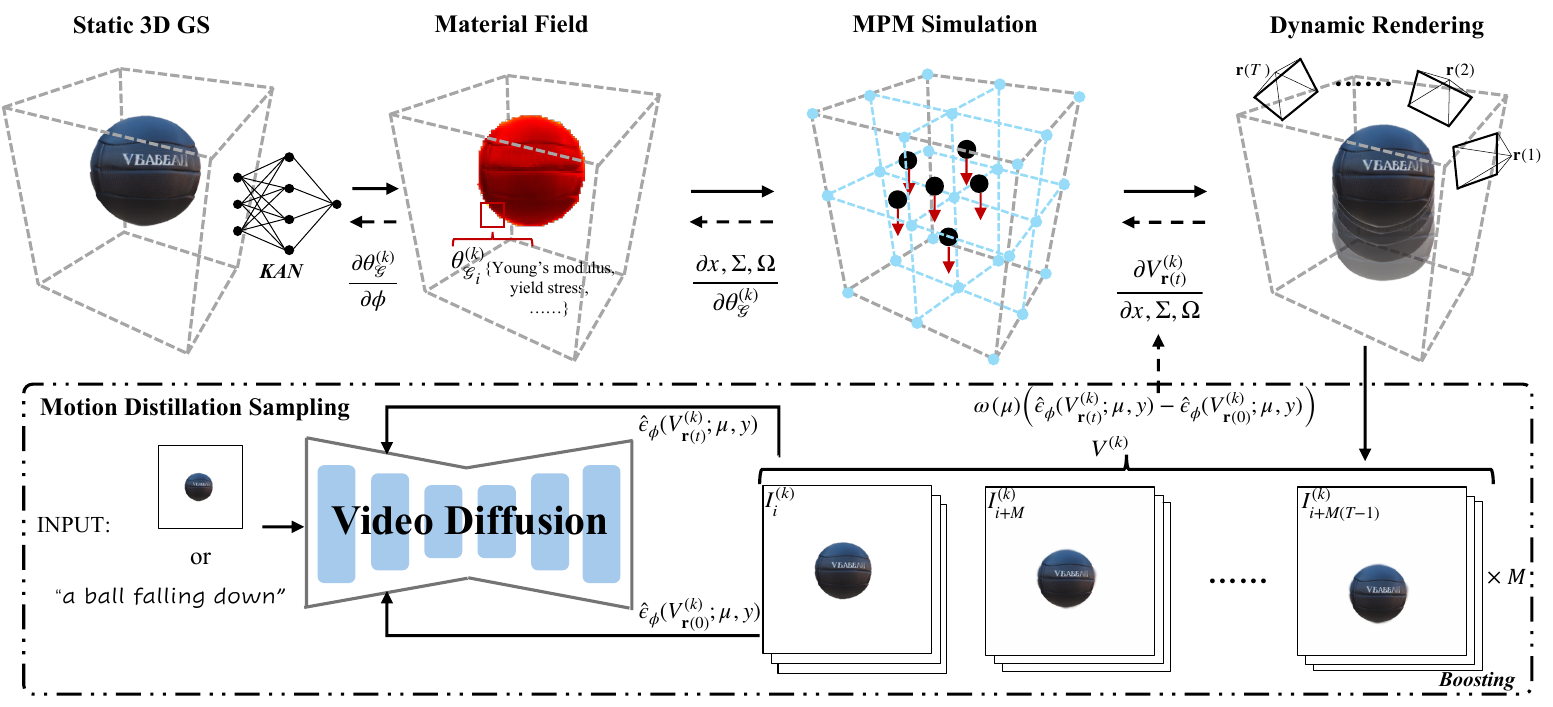}
  \caption{Overview of DreamPhysics. First, a set of physical parameters is initialized with a KAN-based material field for a static 3D GS. Then, it is fed to an MPM simulator to render a 3D video. Finally, we leverage motion distillation sampling to optimize the rendered video, and the distillation gradients are back-propagated to refine the physical parameters.}
  \label{fig:overview}
\end{figure*}

\subsection{Score Distillation Sampling}
The score distillation sampling (SDS)~\cite{poole2022dreamfusion,wang2023score} distills pre-trained 2D diffusion models to the parameters of the 3D representation, widely used in 3D generation methods. Recently, SDS has had various extensions. Variational score Distillation (VSD)~\cite{wang2024prolificdreamer} proposes an additional LoRA term $\bm{\epsilon}_\theta$ to learn the distribution of current 3D scenes, which is attached to the score as:
\begin{equation}\label{eq:vsd}
\nabla_{\theta}\mathcal{L}_{\text{VSD}}(\theta) \triangleq 
\mathbb{E} \left[
    \omega(t)
    \left( \hat{\bm{\epsilon}}_{\text{2D}}(\bm{x}_t,t,y) - \hat{\bm{\epsilon}}_{\theta}(\bm{x}_t,t,c,y)  \right)
    \frac{\partial \bm{x}}{\partial \theta}
\right],
\end{equation}
where $t$ is the noise timestep and $y$ is the input condition. $\hat{\bm{\epsilon}}_{\text{2D}}$ and $\hat{\bm{\epsilon}}_{\theta}$ are noises predicted by a pre-trained 2D diffusion model and the LoRA. Another extension, SDS-T, is for dynamic 3D generation, where video diffusion models are deployed to supervise the time-dependent deformation of static 3D objects. Specifically, given a camera trajectory $\mathbf{r}(t)$, SDS-T optimizes the rendered 3D video $V_{\mathbf{r}(t)}$ with predicted noise $\hat{\epsilon}_{\text{V}}$, as:
\begin{equation}
\label{eq:sdst}
\nabla_{\theta}\mathcal{L}_{\text{SDS-T}}(\theta) \triangleq 
\mathbb{E}\left[
    \omega(\mu)
    \left( \hat{\epsilon}_{\text{V}}(V_{\mathbf{r}(t)};\mu,y) - \epsilon \right)
    \frac{\partial V_{\mathbf{r}(t)}}{\partial \theta}
\right],
\end{equation}
where $\mu$ is noise timestep and $\theta$ is target deformation.

\section{DreamPhysics}
\subsection{Method Overview}
As shown in Figure~\ref{fig:overview}, given a generated object or a reconstructed scene $\{\mathcal{G}_i\}$ represented by 3D GS~\cite{kerbl3Dgaussians}, DreamPhysics aims to estimate the corresponding physical parameters $\{\theta_{\mathcal{G}_i}\}$ for the MPM-based simulator. For each Gaussian kernel $\mathcal{G}_i$, we initialize its parameters $\theta_{\mathcal{G}_i}^{(0)}=\bm{\phi}(x_i)$ with a KAN-based~\cite{liu2024kan} material field $\bm{\phi}$ and then simulate a time-dependent state $\{x_i(t), \Sigma_i(t), \Omega_i(t)\}$, which can be rendered as a $L$-length video $V^{(0)}=\{I_1^{(0)}, I_2^{(0)}, ..., I_L^{(0)}\}$. The rendered video may look unrealistic due to the inaccurate initialization of $\theta_\mathcal{G}^{(0)}$. Therefore, we propose motion distillation sampling (MDS), which distills video diffusion's motion priors while weakening its color bias. The distillation gradient is then propagated backward to the material field, updating corresponding parameters to $\theta^{(1)}_\mathcal{G}$. Similarly, for each training iteration $k$, we can obtain an optimized $\theta^{(k+1)}_\mathcal{G}$ via the distillation of $V^{(k)}$. Considering current video diffusion models' low frame rate, we further propose a frame-boosting strategy to supervise more simulation frames. After several rounds of optimization, the final physical parameters $\hat{\theta}_\mathcal{G}$ can converge to a reasonable range.

\subsection{Parameter Optimization with MDS}
Video generative models are trained with real-world captured videos that cover kinds of physical phenomena. As a result, given a simulated video V, we can assess whether it is natural and realistic based on the judgement of video models. To this end, one direct solution is to treat videos generated by video models as ground truth, supervising $V$ with reconstruction loss~\cite{zhang2024physdreamer}. However, limited by the control capability, existing video generators can hardly produce desired ground-truth videos. We consider exploring distillation methods to optimize simulated results. Motion distillation sample is thus proposed to enhance the distillation of video diffusion's motion priors.


\noindent\textbf{Motion Distillation Sample.}
With the simulation of MPM, a time-dependent state $\{x_i(t), \Sigma_i(t), \Omega_i(t)\}$ is predicted according to Eq.~(\ref{eq:mpm}), representing a motion in the 3D space. Our intention is to optimize this simulated motion. However, the information of a video can be divided into two terms, \textit{i.e.}, color and motion, where color biases between video diffusion models and the simulated video should be dismissed. In VSD~\cite{wang2024prolificdreamer}, a LoRA term pushes the distribution of the target object away from the gradient direction of the current state. Similarly, we can adopt an additional term to omit the information in the color space. We suppose that the first frame can represent the color for a whole video, so our motion distillation sample $s_\text{MDS}$ is formulated as,
\begin{equation}
\label{eq:mds_score}
    \bm{s}_\text{MDS} = \omega(\mu) \left( \hat{\epsilon}_{\text{V}}(V_{\mathbf{r}(t)};\mu,y) - \hat{\epsilon}_{\text{V}}(V_{\mathbf{r}(0)};\mu,y) \right),
\end{equation}
where $\mathbf{r}(0)$ is the camera viewpoint in the first frame.

Note that the gradient of $s_\text{MDS}$ cannot be directly propagated to the target physical parameters $\theta_\mathcal{G}$, and it needs to go through the differentiable MPM. Thus, our training objective can be written as:
\begin{equation}
\label{eq:motion}
\nabla_{\theta_\mathcal{G}}\mathcal{L}_{\text{MDS}}(\theta_\mathcal{G},\mathbf{r}(t)) \triangleq 
\mathbb{E} \left[
    \bm{s}_\text{MDS} 
    \frac{\partial V_{\mathbf{r}(t)}}{\partial x,\Sigma,\Omega}
    \frac{\partial x,\Sigma,\Omega}{\partial \theta_\mathcal{G}}
\right].
\end{equation}

\subsection{Parameter Estimation with Material Field}
The value range for physical properties $\theta_\mathcal{G}$ can be very large, \textit{e.g.}, the reasonable values for Young's modulus can vary from $1e4$ to $1e8$. However, during gradient updates, the same gradient can result in varying update granularity across different magnitudes, causing parameters to get stuck within a specific magnitude range. To enable parameters to converge more quickly to a reasonable range, we propose to perform a KAN-based tri-plane representation to model the material field and conduct frame boosting to further facilitate the training process.


\noindent\textbf{KAN-Based Triplane.}
Tri-plane is widely used to encode spatial information. Given a 3D coordinate $x$, the tri-plane extractor projects it onto three orthogonal planes, \textit{i.e.}, the front view, side view, and top view. These projections match $x$ with 2D features that represent different perspectives of the 3D space. We extract features with KAN~\cite{liu2024kan}, which integrates kernel methods and attention mechanisms to offer superior modeling capabilities for physics-based tasks compared to traditional MLPs. Extracted features are then combined to form a unified representation, which constitutes our physical parameters $\theta_\mathcal{G}$. The gradient is propagated as:
\begin{equation}
    \nabla_{\bm{\phi}}\mathcal{L}_{\bm{\phi}}(x, \mathbf{r}(t)) \triangleq 
    \mathbb{E} \left[
    \mathcal{L}_{\text{MDS}}(\bm{\phi}(x),\mathbf{r}(t)) \frac{\partial \theta_\mathcal{G}}{\partial \bm{\phi}}
    \right].
\end{equation}

\noindent\textbf{Frame Boosting.}
The MPM simulator is a sequential model, which can easily lead to gradient vanishing or exploding like RNN~\cite{rumelhart1986learning}. We have to conduct truncated back-propagation through time (BPTT), preserving the gradient of key frame simulation only. Truncated BPTT can effectively prevent gradient issues, but the supervision could be limited to specific frames.
To ensure that our supervision covers as many video frames as possible, we further suggest a frame-boosting strategy. Specifically, given a total number of frames $M\times T$, we can separate them into $M$ groups of frames with equal intervals, \textit{i.e.}, $V_{t_i}=\{I_i, I_{i+M}, ..., I_{i+M(T-1)}\}$ for the $i$-th group. These groups formulate different videos, which are fed into the supervision process alternately. Finally, the boosted motion distillation can be formulated as:
\begin{equation}
\mathcal{L}_{\hat{\bm{\phi}}}(x) = \frac{1}{M}\sum_{i=1}^M \mathcal{L}_{\bm{\phi}}(x, \mathbf{r}(t_i)),
\end{equation}
where $\hat{\bm{\phi}}$ is the boosted material field.

\section{Experiments}
In this section, we show our 4D generation content on both text-conditioned and image-conditioned optimizations and compare it with previous state-of-the-art methods. Extensive ablation studies are then conducted to demonstrate the effectiveness of our newly proposed components.

\subsection{Experimental Setup}
\noindent\textbf{Implementation Details.} The simulation is based on the warp~\cite{warp2022} implementation of MPM~\cite{stomakhin2013material,jiang2016material}. For most simulation scenes, we set the simulation duration as $5\times10^{-5}$ second and the frame duration as $4\times10^{-2}$ second. Thus, we simulate 800 steps between every two renderings and include the simulation gradient of the last step in the optimization. We leverage a text-to-video diffusion model ModelScope~\cite{wang2023modelscope} and an image-to-video diffusion model Stable Video Diffusion (SVD)~\cite{blattmann2023stable} to conduct text-conditioned and image-conditioned optimization, respectively. The numbers of their generated video frames $T$ are 16 and 25, respectively. For frame boosting, we set $M=5$, boosting the video slices to 5 groups. The setting of MDS follows SDS, where CFG value is set to 100. We stop the training if optimized parameter values stabilize within one order of magnitude. The training process requires around 30 iterations. The iteration time highly depends on the number of input Gaussian kernels, and it is within 30 seconds for most cases.


\noindent\textbf{Dataset.} We collect seven 3D static scenes or objects from previous works~\cite{xie2023physgaussian,zhang2024physdreamer} and 3D GS generative models~\cite{tang2024lgm}. The content includes three plants, a beanie hat, a telephone cord, a sofa with pillows, and a ball, where two motions (rotation and collision) are involved in the simulator.


\noindent\textbf{Evaluation Metric.} We use the aesthetic quality from VBench~\cite{huang2023vbench}, grading the artistic score from 0 to 10 using the LAION aesthetic predictor~\cite{laion2022aesthetic}. This metric can reflect aesthetic aspects such as the naturalness of the video, which exactly meets our evaluation requirements.
In addition, we will add user study results in the supplementary materials.


\noindent\textbf{Compared Methods.} Since physics-based 4D generation is still under development, we compare three existing methods PhysGaussian~\cite{xie2023physgaussian}, PhysDreamer~\cite{zhang2024physdreamer}, and DreamGaussian4D~\cite{ren2023dreamgaussian4d}. PhysGaussian is a pioneer work that manually sets all the physical properties in a physics-based simulator. PhysDreamer is a concurrent work that supervises physical parameters with ground-truth videos. DreamGaussian4D predicts the deformation of 3D GS without physical constraint, which is different from the above two works.

\begin{figure*}[t]
  \centering
  \includegraphics[width=0.85\textwidth]{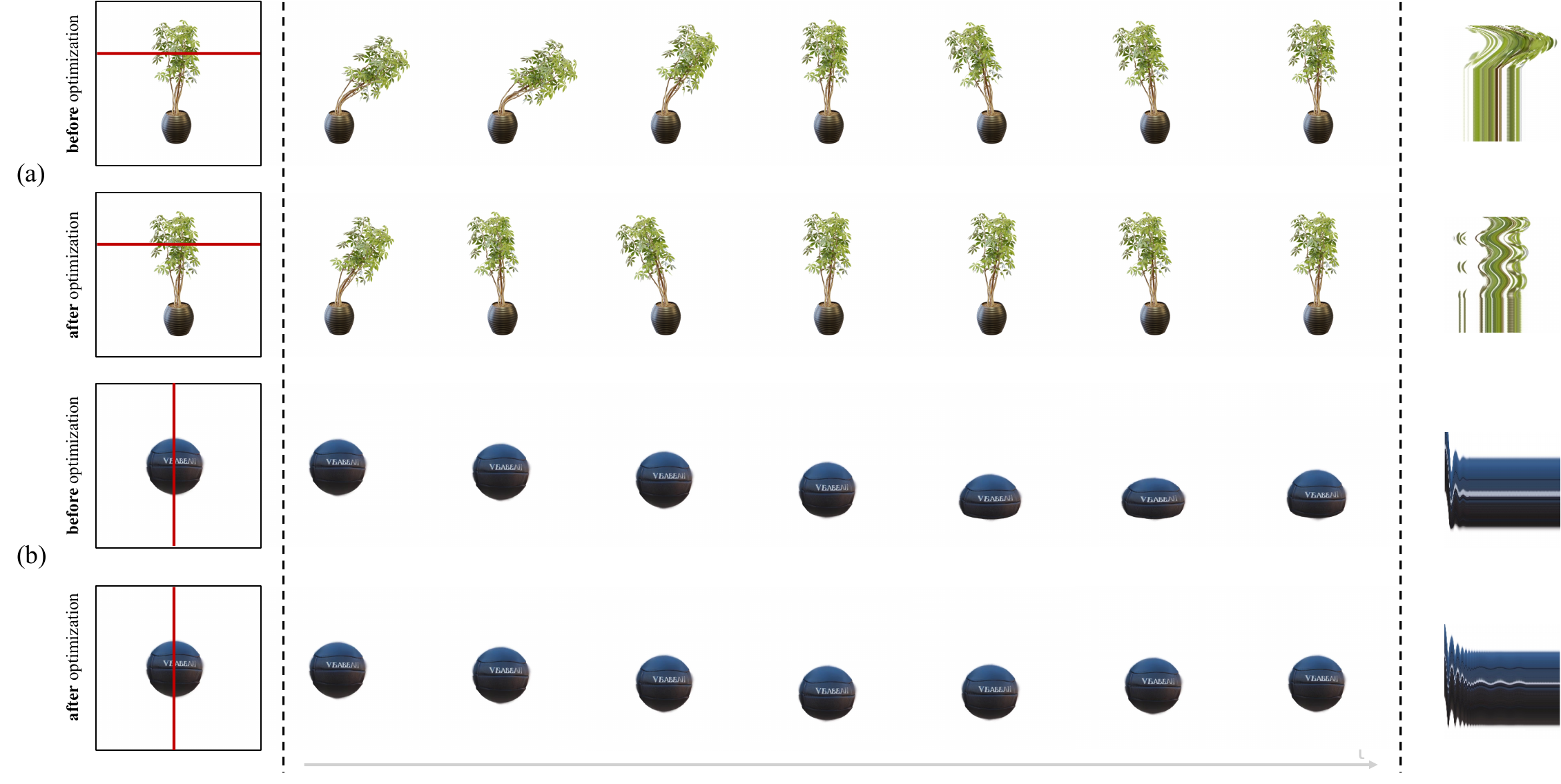}
  \caption{(a) Text-conditioned optimization; (b) Image-conditioned optimization.
  Right images are the space-time (X-t) slices, one axis represents time and the other axis shows a space slice (red line) of the object.}
  \label{fig:vis1}
\end{figure*}

\subsection{3D Dynamics Generation}
\noindent\textbf{Text Condition.}
In Figure~\ref{fig:vis1}(a), we select the ficus scene in PhysGaussian~\cite{xie2023physgaussian} and input a text prompt "\textit{ficus swaying in the wind}" to simulate the rotation motion. The ficus would excessively tilt to one side and have difficulty returning to its original position if its Young's modulus is set too low. After the optimization by our DreamPhysics, Young's modulus falls within a normal range, and the swaying looks more natural. From the space-time slices, the optimized motion trajectory looks more realistic.

\begin{figure*}[t]
  \centering
  \includegraphics[width=0.87\textwidth]{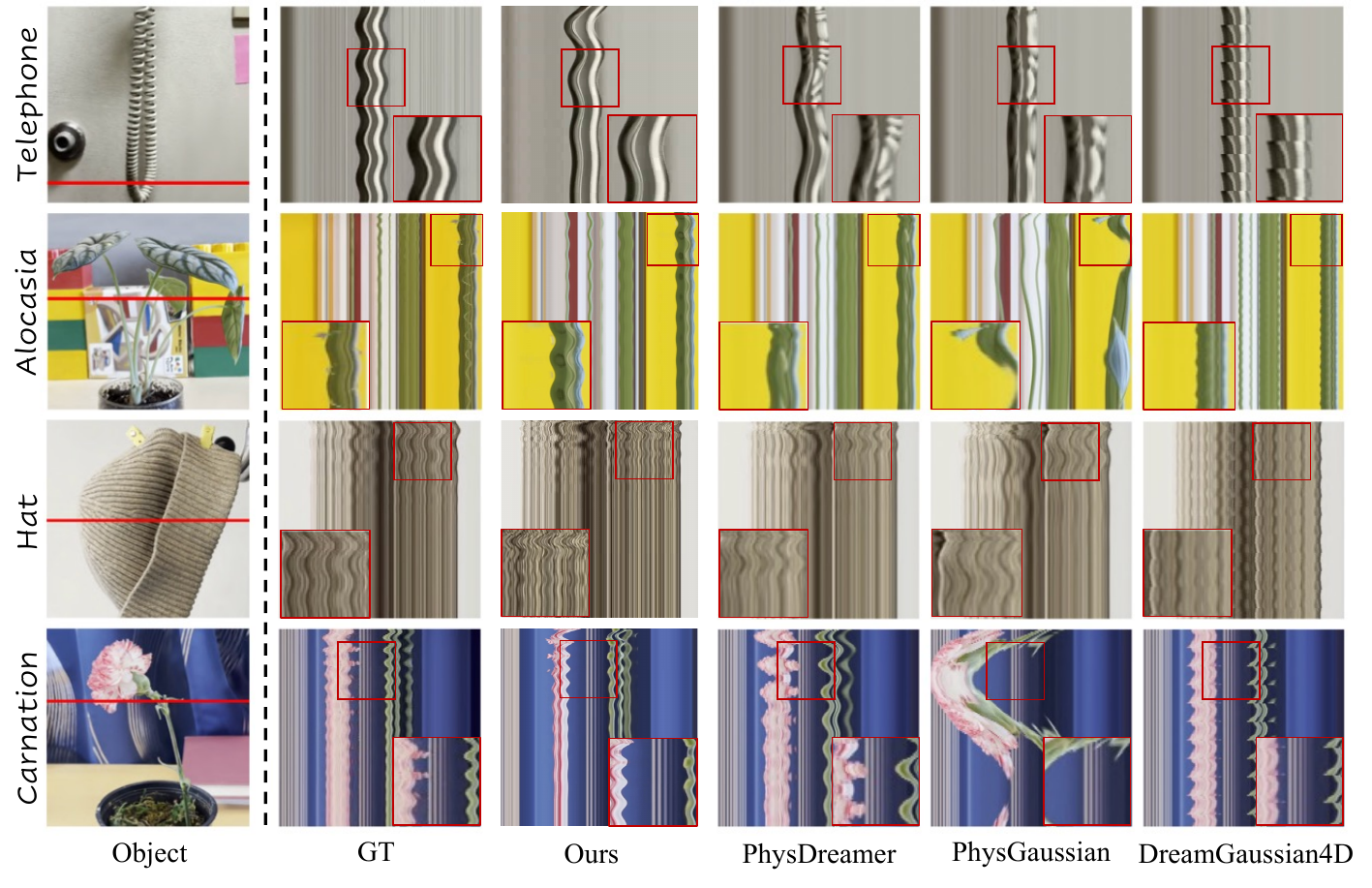}
  \caption{Viualization of space-time slices. Compared with previous works, our results are more close to the ground truth.}
  \label{fig:vis2}
\end{figure*}

\begin{figure*}[!htbp]
  \centering
  \includegraphics[width=0.85\textwidth]{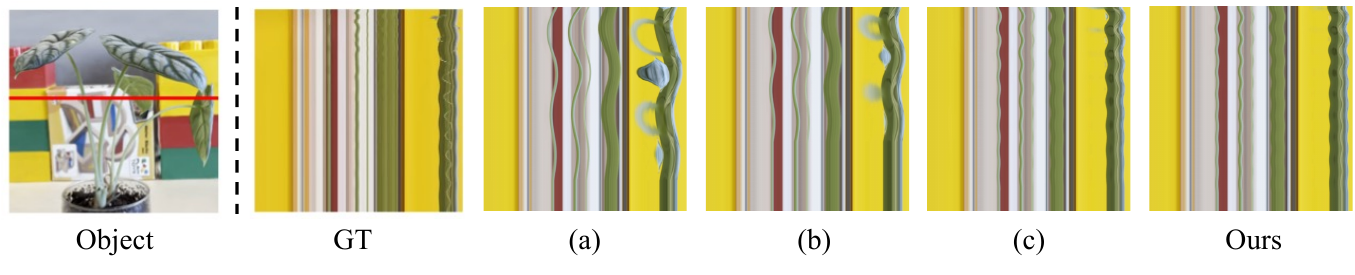}
  \caption{Visualization of space-time slices for ablation study. (a) and (b) are not quite consistent with the ground truth. (c) and our method can generate closer content compared with the ground truth.}
  \label{fig:ab}
\end{figure*}


\noindent\textbf{Image Condition.}
For image-conditioned optimization, the first frame is regarded as the input image. We select a generated ball and try to optimize its dropping process, which is an example of collision motion, as shown in Figure~\ref{fig:vis1}(b). When hitting the ground, the ball would exhibit excessive deformation if the physical properties are not initialized accurately. Our method can effectively adjust these properties to a reasonable range after the optimization.

\begin{table}[t]
    \centering
    \setlength\tabcolsep{0.85mm} 
    \caption{Quantitative results for the comparison with previous works on 4 scenes from Figure~\ref{fig:vis2}. The higher aesthetic quality score indicates better generation quality.}
    \scalebox{0.95}
    {
    \begin{tabular}{cccc|c}
        \hline
        DreamGaussian4D & PhysGaussian & PhysDreamer  &  Ours & GT \\
        \hline
        4.61  & 4.98 & 4.84  & \textbf{5.03} & 5.13  \\
        \hline
    \end{tabular}
    }
    \label{tab:comparison}
\end{table}

\noindent\textbf{Comparison with State-of-the-art Works.}
We report the quantitative results of all the compared methods in Table~\ref{tab:comparison}. Since PhysDreamer hasn't released its training implementation, we can only compare four evaluation scenes, where the corresponding ground-truth videos are provided in the video demo. Considering that other methods don't have extra text inputs, we use the first frame as the image condition to conduct the optimization. According to the evaluation of aesthetic quality, our results are the closest to the ground truth. PhysDreamer has a lower score compared with PhysGaussian, which indicates that pre-generated videos may not be a proper ground truth for supervision. The generation quality of DreamGaussian4D is the worst because its deformation prediction didn't consider physical constraints. 

We also provide the visualization of space-time slices in Figure~\ref{fig:vis2}. Since all the physical properties in PhysGaussian are manually set, its generated motions often look too extreme. DreamGaussian4D generates the most consistent motions but appears less natural, as its prediction lacks physical constraint. PhysDreamer can exhibit energy dissipation to some extent, while our results look more similar to the ground-truth visualization, in terms of amplitude and frequency of the simulated motions.

\begin{table}[t]
    \centering
    \caption{Quantitative results of ablation study on 7 scenes. \textit{\textbf{Score}} denotes the average aesthetic quality score, and \textit{\textbf{Iter}} denotes the average training iterations.}
    \begin{tabular}{c|ccc|cc}
        \hline
        Method\hspace{-1.5mm} & +KAN & +$\mathcal{L}_\text{MDS}$ & +Boost & \textit{\textbf{Score}}$\uparrow$ & \textit{\textbf{Iter}}$\downarrow$ \\
        \hline
         (a)\hspace{-1.5mm} & & & & 4.86 & 36.86 \\
         (b)\hspace{-1.5mm} & \checkmark & & & 4.89 & 34.29 \\
         (c)\hspace{-1.5mm} & \checkmark & \checkmark & & \textbf{4.94} & 33.86 \\
         Ours\hspace{-1.5mm} & \checkmark & \checkmark & \checkmark & 4.93 & \textbf{29.71} \\
        \hline
    \end{tabular}
    \label{tab:ab}
\end{table}

\subsection{Ablation Study}
To evaluate the effectiveness of our newly proposed modules, we conduct ablation studies on all 7 scenes. Our baseline uses a vanilla SDS-T loss (Eq.~(\ref{eq:sdst})), where gradients are propagated to the physical parameters without KAN. Based on this, we attach our KAN-based material field, motion distillation sampling, and frame boosting step by step. 

We report the aesthetic quality score and training iterations in Table~\ref{tab:ab}. In (a), the physical parameters can hardly converge to a reasonable range, with the evaluation score and required iterations being the worst. Equipped with a KAN-based material field, (b) can facilitate the optimization and improve the generation quality. Then, we use motion distillation sampling $\mathcal{L}_\text{MDS}$ in (c), where the aesthetic score is further improved. In (d), our final method enjoys a faster optimization speed within 30 training iterations, demonstrating that our frame boosting can fasten the parameter convergence. Note that, frame boosting is not designed for optimization quality, so our final score is similar to (c).

We provide the visualization of the Alocasia scene in Figure~\ref{fig:ab}. The space-time slices of (a) and (b) are not quite consistent with the ground truth, while (c) and our final method can produce 4D content that is competitive to real-captured videos. These results are consistent with our quantitative results in Table~\ref{tab:ab}.

\section{Conclusion}
In this work, we introduced a new framework DreamPhysics, which learns the physical properties of 3D Gaussian Splatting with video diffusion priors. Based on the physics-based simulation, DreamPhysics distills the motion priors to physical parameters with motion distillation sampling. To facilitate that process, we further propose a KAN-based material field with frame boosting. Extensive experiments demonstrate that our method can produce high-quality 4D content with both text and image conditions.

Albeit the improvement compared with previous works, the physics-based 3D dynamics research still faces two problems, \textit{i.e.}, simulated motions and scene-level interaction. Each kind of motion depends on independent physical constraints. Current frameworks can hardly combine all the motions into one simulator. Moreover, simulators can only handle the interactions of a few target objects, but environments are dismissed. For example, in the simulation of the telephone (Figure~\ref{fig:vis2}), shadows on the wall cannot change with the movement of the telephone cord. We will explore these problems for future work.

\section{Acknowledgments}
This work is in part supported by the National Key R\&D Program of China (2021YFF0900500), the National Natural Science Foundation of China (NSFC) under grants 62441202, and two GRF grants from the Research Grants Council of Hong Kong (RGC No.: 11211223 and 11220724).

\bibliography{aaai25}

\clearpage
\setcounter{page}{1}

\section*{Additional Experiments}
\subsection*{User Study}
To better assess human preferences for generated 3D videos, we conducted user studies in both SOTA comparisons and ablation experiments. For each scenario, we provided four video clips and asked the participants to select the most preferred one. The selection criteria are the realism and coherence of the generated videos. A total of 28 volunteers participated in the study, including 5 professionals from the 3D art industry.

From Table~\ref{tab:user}, our method is the most favorable one in both SOTA comparisons and ablation experiments. The results are generally consistent with the quantitative evaluation metric, \ie, aesthetic quality used in the main paper.

\begin{table*}[t]
    \centering
    \caption{User studies on the comparison of state-of-the-art methods and ablation methods.}
    \begin{tabular}{c|ccccccc|c}
        \hline
        Method & \textit{\textbf{Alocasia}} & \textit{\textbf{Carnation}} & \textit{\textbf{Hat}} & \textit{\textbf{Telephone}} & \textit{\textbf{Ball}} & \textit{\textbf{Ficus}} & \textit{\textbf{Pillow}} & Total \\
        \hline
         DreamGaussian4D & $3.57\%$ & $7.14\%$ & $10.71\%$ & $7.14\%$ & - & - & - & $7.14\%$ \\
         PhysGaussian & $21.43\%$ & $10.71\%$ & $28.57\%$ & $3.57\%$ & - & - & - & $16.07\%$ \\
         PhysDreamer & $17.86\%$ & $35.71\%$ & $7.14\%$ & $25.00\%$ & - & - & - & $21.43\%$ \\
         Ours & $\bm{57.14\%}$ & $\bm{46.43\%}$ & $\bm{53.57\%}$ & $\bm{64.29\%}$ & - & - & - & $\bm{55.36\%}$ \\
         \hline
         (a) & $0\%$ & $7.14\%$ & $21.43\%$ & $10.71\%$ & $0\%$ & $0\%$ & $25\%$ & $9.18\%$ \\
         (b) & $7.14\%$ & $21.43\%$ & $7.14\%$ & $7.14\%$ & $\bm{35.71}\%$ & $17.86\%$ & $7.14\%$ & $14.80\%$ \\
         (c) & $35.71\%$ & $25.00\%$ & $28.57\%$ & $32.14\%$ & $32.14\%$ & $35.71\%$ & $17.86\%$ & $29.59\%$ \\
         Ours & $\bm{57.14\%}$ & $\bm{46.43\%}$ & $\bm{42.86\%}$ & $\bm{50.00\%}$ & $32.14\%$ & $\bm{46.43}\%$ & $\bm{50.00}\%$ & $\bm{46.43\%}$ \\
        \hline
    \end{tabular}
    \label{tab:user}
\end{table*}

\begin{figure*}[!htbp]
  \centering
  \includegraphics[width=0.99\textwidth]{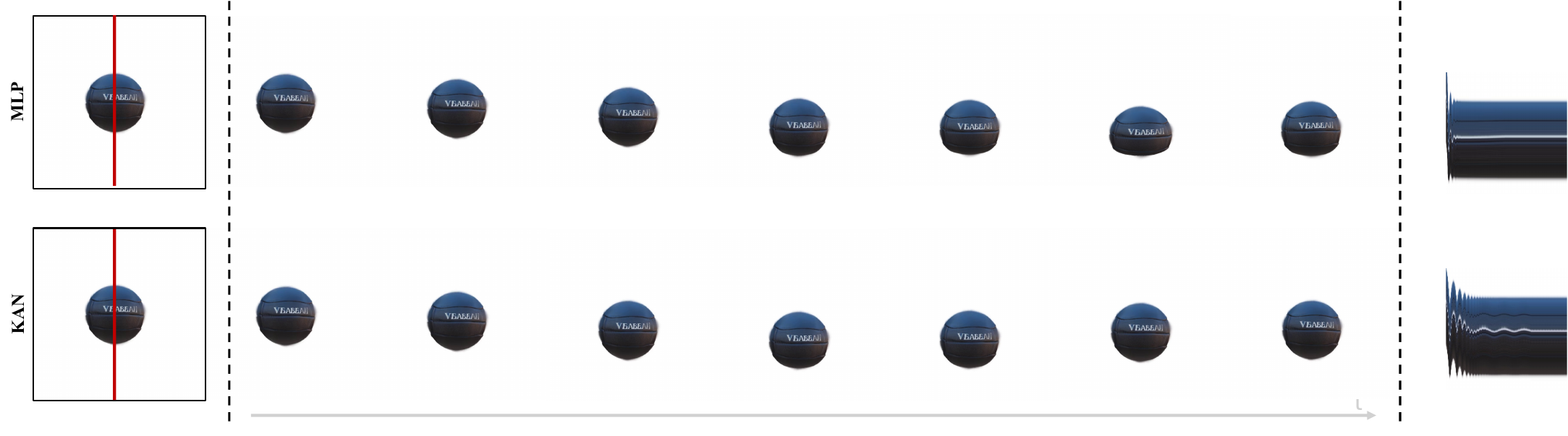}
  \caption{Visualization of ablation study on KAN.}
  \label{fig:ball}
\end{figure*}

\subsection*{Ablation Study on KAN}
In the tri-plane of our material field, we replace the MLP encoder with KAN~\cite{liu2024kan} layers to enhance the modeling of physical parameters. We conduct an ablation study on KAN layers based on our final method, using a classic MLP encoder to extract tri-plane features, instead of KAN layers.
The results show that MLP generally encounters some issues in collision scenarios. As shown in Figure~\ref{fig:ball}, the ball optimized by MLP is too soft to maintain its original shape, while our method can restore it.

\subsection*{Video Visualization Results}
We provide generated videos in the supplementary materials. Please refer to the HTML file named ``results.html'' or directly check the MP4 files in the ``videos'' folder.

\section*{Experimental Details}
\subsection*{Dataset Scenes}
Seven scenes are used in our experiments. Their details are as follows.

\noindent\textbf{Alocasia.} This scene is from PhysDreamer~\cite{zhang2024physdreamer}, showing an alocasia swaying on a table. The simulated motion is the rotation of an elastic object. This scene has a real-captured video as the ground truth.

\noindent\textbf{Carnation.} This scene is from PhysDreamer, showing a carnation swaying on a table. The simulated motion is the rotation of an elastic object. This scene has a real-captured video as the ground truth.

\noindent\textbf{Hat.} This scene is from PhysDreamer, where a hat hanging on a clothes hanger is swaying. The simulated motion is the rotation of an elastic object. This scene has a real-captured video as the ground truth.

\noindent\textbf{Telephone.} This scene is from PhysDreamer, showing a telephone on the wall, with its cord swaying. The simulated motion is the rotation of an elastic object. This scene has a real-captured video as the ground truth.

\noindent\textbf{Ball.} Since all the evaluation scenes in PhysDreamer are used to simulate the rotation of elastic objects, we propose a collision example where a ball drops to the ground. The ball is generated by LGM~\cite{tang2024lgm}. We place it in a gravitational field.

\noindent\textbf{Ficus.} This scene is from PhysGaussian~\cite{xie2023physgaussian}, showing a ficus swaying in the wind. The simulated motion is the rotation of an elastic object.

\noindent\textbf{Pillow.} This scene is from PhysGaussian~\cite{xie2023physgaussian}, where a cushion and three pillows fall onto the sofa one after another. This simulated motion is a complicated collision motion.

\subsection*{Evaluation Metrics}
Traditional image evaluation metrics like SSIM~\cite{wang2004image}, PSNR~\cite{huynh2008scope}, LPIPS~\cite{zhang2018unreasonable}, and FID~\cite{heusel2017gans} measure the similarity of two input images, which can hardly evaluate the motion consistency compared with the ground truth. FVD~\cite{unterthiner2019fvd} takes temporal information into consideration, extracting spatio-temporal features with I3D~\cite{carreira2017quo}. However, similarly to FID, this metric also focuses on frame content, rather than motion fidelity.

A recent video benchmark VBench~\cite{huang2023vbench} proposes to evaluate the temporal quality of video generation with motion smoothness. In the evaluation of motion smoothness, odd-number frames are dropped from input videos and then predicted by a frame interpolation model~\cite{li2023amt}. The smoothness score is related to the similarity of the original frames and predicted frames. Although its motivation is to measure the smoothness of generated motion, this metric can hardly discriminate the quality of physics-based simulation results. All the physics-based methods (PhysGaussian, PhysDreamer, and our method) can achieve over 0.99 in this metric because the simulator is capable of producing continuous motions.

Actually, our evaluation objective should be judging whether a generated motion conforms to real-world physical laws. In other words, we should assess whether the generated motion looks realistic. As a result, we use aesthetic quality in VBench, calculating the average aesthetic score of all the frames as our evaluation metric. In this way, some of the frames can get low scores if the physical parameters are set inappropriately.
Furthermore, the user study can better evaluate the realism and coherence of the generated videos. We also perform user studies in Table~\ref{tab:user}.

\subsection*{Code and Pseudocode}
We provide codes in the supplementary materials. Please refer to the ``code'' folder. Here, we further provide a pseudocode in Algorithm~\ref{alg:ours}.

\begin{algorithm}[t] 
    \centering 
    \small
    \caption{DreamPhysics}
    \label{alg:ours} 
    \begin{algorithmic}[1] 
        \STATE\textbf{Input}: text/image condition $y$ and static 3D GS scene $\{\mathcal{G}_i\}$
        \STATE \textbf{Initialize} material field $\phi^{(0)}$ and MPM simulator $\mathcal{M}$
        \STATE\textbf{Load}: video diffusion model v
        \WHILE{not \textit{converged}}
            \STATE $\blacktriangleright$ \textbf{Step 1: Physics-Based Simulation}
            \STATE Extract physical parameters $\{\theta_{\mathcal{G}_i}\}$ as $\theta_{\mathcal{G}_i}=\phi^{(k)}(x_i)$
            \STATE Simulate time status $\mathcal{M}(\mathcal{G},\theta)=\{x_i(t), \Sigma_i(t), \Omega_i(t)\}$
            \STATE $\blacktriangleright$ \textbf{Step 2: Motion Distillation Sampling}
            \STATE Sample viewpoints $\{c_1,c_2,...,c_\text{MT}\}$ and timestep $\mu$
            \STATE Render video frames $\{I_1,I_2,...,I_\text{MT}\}$ and split M groups of videos $\{V_{\mathbf{r}_{(ti)}}=[I_i,I_{i+M},...,I_{i+M(T-1)}]\}$, $i=1,...,M$
            \STATE $\bm{s}_\text{MDS} = \omega(\mu) \left( \hat{\epsilon}_{\text{V}}(V_{\mathbf{r}(t_i)};\mu,y) - \hat{\epsilon}_{\text{V}}(I_i;\mu,y) \right)$
            \STATE $\nabla_{\theta_\mathcal{G}}\mathcal{L}_{\text{MDS}}(\theta_\mathcal{G},\mathbf{r}(t_i))\triangleq 
\mathbb{E} \left[
    \bm{s}_\text{MDS} 
    \frac{\partial V_{\mathbf{r}(t)}}{\partial x,\Sigma,\Omega}
    \frac{\partial x,\Sigma,\Omega}{\partial \theta_\mathcal{G}}
\right]$
            \STATE $\blacktriangleright$ \textbf{Step 3: Gradient Propagation}
            \STATE $\nabla_{\bm{\phi}}\mathcal{L}_{\bm{\phi}}(x, \mathbf{r}(t)) \triangleq 
    \mathbb{E} \left[
    \mathcal{L}_{\text{MDS}}(\phi^{(k)}(x),\mathbf{r}(t)) \frac{\partial \theta_\mathcal{G}}{\partial \bm{\phi}}
    \right]$
            \STATE $\phi^{(k+1)} \gets \phi^{(k)} - \nabla_\phi \frac{1}{M} \sum_{i=1}^M \mathcal{L}_\phi(x,\mathbf{r}(t_i))$
            \STATE $\blacktriangleright$ \textbf{Check Convergence}
            \IF{$[\phi^{(k+1)}$, $\phi^{(k)}$, $\phi^{(k-1)}]$ in same order of magnitude}
                \STATE \textit{converged} $\gets$ \textbf{True}
            \ENDIF
        \ENDWHILE
        \RETURN physical parameters $\theta_{\mathcal{G}_i}=\phi^{(k+1)}(x_i)$
    \end{algorithmic}
\end{algorithm}

\section*{Theoretical Details}
\subsection*{Material Point Method}
The material point method (MPM) is a powerful numerical technique used to simulate the behavior of continuum materials. MPM discretizes a material body into a collection of material points (often referred to as particles), each carrying properties such as mass, velocity, deformation gradient, and stress. These particles are coupled with a background computational grid, which aids in the calculation of spatial derivatives and the application of external forces.

MPM operates through two key phases: Particle-to-Grid (P2G) Transfer and Grid-to-Particle (G2P) Transfer.

\noindent\textbf{Particle-to-Grid (P2G) Transfer.}
In this phase, the mass and momentum of particles are transferred to the grid nodes using interpolation functions. The mass at a grid node \( i \) is computed as:
\[
m_i^n = \sum_p w_{ip}^n m_p,
\]
where \( m_p \) is the mass of particle \( p \), and \( w_{ip}^n \) is the interpolation weight (often derived from a B-spline kernel) between particle \( p \) and grid node \( i \). The momentum at the grid node is similarly updated:
\[
m_i^n \mathbf{v}_i^n = \sum_p w_{ip}^n m_p \left( \mathbf{v}_p^n + \mathbf{C}_p^n (\mathbf{x}_i - \mathbf{x}_p^n) \right),
\]
where \( \mathbf{v}_p^n \) is the velocity of particle \( p \), \( \mathbf{C}_p^n \) represents the affine velocity field gradient, and \( \mathbf{x}_i \) and \( \mathbf{x}_p^n \) are the positions of the grid node and particle, respectively.

The grid velocities are then updated based on the external forces and internal stresses computed from the particle data:
\[
\mathbf{v}_i^{n+1} = \mathbf{v}_i^n - \frac{\Delta t}{m_i^n} \sum_p \boldsymbol{\tau}_p^n \nabla w_{ip}^n V_p^0 + \Delta t \mathbf{g},
\]
where \( \Delta t \) is the time step, \( \boldsymbol{\tau}_p^n \) is the stress tensor of the particle \( p \), \( V_p^0 \) is the initial volume of the particle, and \( \mathbf{g} \) is the acceleration due to gravity.

\noindent\textbf{Grid-to-Particle (G2P) Transfer.}
After updating the grid, the changes in velocity and momentum are transferred back to the particles. The velocity of particle \( p \) is updated as:
\[
\mathbf{v}_p^{n+1} = \sum_i \mathbf{v}_i^{n+1} w_{ip}^n,
\]
and the new position of the particle is given by:
\[
\mathbf{x}_p^{n+1} = \mathbf{x}_p^n + \Delta t \mathbf{v}_p^{n+1}.
\]
Additionally, the affine velocity field gradient \( \mathbf{C}_p^{n+1} \) and deformation gradient \( \mathbf{F}_p^{n+1} \) are updated as:
\[
\mathbf{C}_p^{n+1} = \frac{4}{(\Delta x)^2} \sum_i w_{ip}^n \mathbf{v}_i^{n+1} (\mathbf{x}_i - \mathbf{x}_p^n)^T,
\]
\[
\mathbf{F}_p^{n+1} = (\mathbf{I} + \Delta t \mathbf{C}_p^{n+1}) \mathbf{F}_p^n.
\]

MPM combines the advantages of Lagrangian (particle-based) and Eulerian (grid-based) methods, making it particularly effective for simulating materials that undergo large deformations, fractures, and complex interactions. It has been successfully applied to a variety of materials, including solids, fluids, granular media, and textiles. Moreover, its suitability for parallel computation on GPUs enables high-performance simulations of large-scale problems.

\subsection*{Score Distillation Sampling}
Score Distillation Sampling (SDS) is a core technique introduced in DreamFusion~\cite{poole2022dreamfusion}. The method is a significant advancement in the realm of generating 3D content by 2D diffusion models. Its goal is to create 3D objects that, when rendered from various angles, look like realistic images. Traditional diffusion models are typically used to generate outputs that match the dimensionality of their training data (e.g., 2D images). However, the challenge here is to leverage these models to optimize 3D structures.

To bridge the gap between 2D diffusion models and 3D object creation, DreamFusion uses Differentiable Image Parameterization (DIP). In this approach, a differentiable generator \( g \) transforms a set of parameters \( \theta \) into an image \( \mathbf{x} = g(\theta) \). For 3D model creation, \( \theta \) represents the parameters of a 3D volume, and \( g \) is a volumetric renderer that generates 2D images from different viewpoints.

SDS optimizes the 3D parameters \( \theta \) so that the generated image \( \mathbf{x} = g(\theta) \) appears like a sample from a pre-trained, frozen diffusion model. The key idea is to bypass the expensive computation of the full diffusion model gradient by simplifying the process.

The gradient used for optimizing \( \theta \) in SDS is derived as:
\begin{align}
\nabla_{\theta} \mathcal{L}_{\text{SDS}}(\mathbf{x}=g(\theta)) \triangleq \mathbb{E}_{t, \epsilon}\left[w(t)\left(\hat{\epsilon}_{\text{2D}}(\mathbf{z}_t; y, t)  - \epsilon\right) \frac{\partial \mathbf{x}}{\partial \theta}\right],
\end{align}
where:
\begin{itemize}
    \item \( \hat{\epsilon}_{\text{2D}}(\mathbf{z}_t; y, t) \) is the predicted noise by the 2D diffusion model at time step \( t \).
    \item \( \epsilon \) is the actual noise added.
    \item \( \frac{\partial \mathbf{x}}{\partial \theta} \) is the Jacobian of the image with respect to the 3D parameters.
\end{itemize}

This gradient effectively guides the 3D model parameters to generate images that align more closely with the high-density regions (plausible images) defined by the diffusion model.

SDS is a groundbreaking method that repurposes 2D diffusion models to guide the creation of 3D models. By optimizing a differentiable parameterization of a 3D volume, SDS allows the generation of complex 3D structures that, when rendered, produce images consistent with the output of the original 2D diffusion model. This approach significantly broadens the applicability of diffusion models beyond their traditional 2D domain, enabling the efficient creation of detailed and realistic 3D models.

\end{document}